\documentclass{article}
\usepackage{spconf}
\usepackage{algorithm,graphicx}
\usepackage{algorithmic}
\usepackage{url}
\usepackage{cite}
\usepackage[usenames]{color}
\usepackage{amsfonts}
\usepackage{fancyhdr}
\usepackage{listings}%
\usepackage{arydshln}
\usepackage{amsmath,amssymb,amsthm}%
\usepackage{graphicx,psfrag,caption,subcaption}
\usepackage{listings}%
\usepackage{arydshln}
\usepackage{subfig}%
\usepackage{algorithmic}%
\newtheorem{myprob}{Problem}%
\newtheorem{myrem}{Remark}%

\renewcommand{\thefootnote}{\fnsymbol{footnote}}

\DeclareMathOperator*{\argmin}{arg\,min}

\title{Learning Sparse Graphs Under Smoothness Prior}
\name{Sundeep Prabhakar Chepuri$^\dag$, Sijia Liu$^\ddag$, Geert Leus$^\dag$, and Alfred O. Hero III$^\ddag$ \thanks{This work  is supported in part by the KAUST-MIT-TUD consortium under grant~{OSR-2015-Sensors-2700} and the US Army Research Office under grant W911NF-15-1-0479.}}
\address{$^\dag$Delft University of Technology (TU Delft), The Netherlands\\
$^\ddag$Department of EECS, University of Michigan, Ann Arbor, MI 48109,  USA\\
Email:~$^\dag$\{s.p.chepuri; g.j.t.leus\}@tudelft.nl, $^\ddag$\{lsjxjtu, hero\}@umich.edu.}

\begin{document}
\ninept
\maketitle
\begin{abstract}
In this paper, we are interested in learning the underlying graph structure behind training data. Solving this basic problem is essential to carry out any graph signal processing or machine learning task. To realize this, we assume that the data is smooth with respect to the graph topology, and we parameterize the graph topology using an edge sampling function. That is, the graph Laplacian is expressed in terms of a sparse edge selection vector, which provides an explicit handle to control the sparsity level of the graph. We solve the sparse graph learning problem given some training data in both the noiseless and noisy settings. Given the true smooth data, the posed sparse graph learning problem can be solved optimally and is based on simple rank ordering. Given the noisy data, we show that the joint sparse graph learning and denoising problem can be simplified to designing only the sparse edge selection vector, which can be solved using convex optimization. 
\end{abstract}
\vspace*{-1mm}
\maketitle

\section{Introduction}

Graphs offer a way to describe and explain relationships in complex datasets, a central entity of modern data analysis, where data 
deluge is prominent~\cite{shuman2013Emerging,sandryhaila2014big,slavakis2014modeling}. In particular, the nodes of the graph denote the entities and the edges 
encode the pairwise relationship between these entities. Such entities are referred to as \emph{graph signals}. Examples of such complex-structured data beyond traditional time-series include data residing on brain networks,  gene networks, social networks, recommendation systems, transportation networks, and so on. 

Having a good quality graph is central to any graph signal processing or machine learning task. In this paper, we are interested in the problem of learning the hidden graph topology behind the data. Due to the sheer quantity of data, we are motivated to select the simplest graphical models that adequately explain the data. In particular, we are interested in learning a sparse graph, i.e., a graph with a limited number of edges that adequately explains the input (or training) data. To realize this, we make a simple, but widely used assumption~\cite{shuman2013Emerging,dong2014learning} that the data is smooth with respect to the discovered graph. 

The contributions in this paper are threefold. First, we model the graph learning problem as an edge selection problem, where we parameterize the graph through a \emph{sparse edge sampling vector}. In particular, the proposed model provides an elegant handle to control the number of edges, thus the graph sparsity. Second, for the case when the true smooth graph signals are given, the graph learning problem can be solved optimally, and the solution is based on simple \emph{rank ordering}. Finally, given the noisy graph signals, i.e., for the joint sparse graph learning and denoising problem, we provide a one-step solution based on convex optimization as well as an algorithm based on alternating minimization.   

The problem of learning the graph Laplacian or the weighted adjacency matrix from smooth graph signals has been considered before~\cite{dong2014learning,kalofolias2016learn}. Learning sparse graphs from the true graph signals, which is the problem we consider in Section~\ref{sec:Noiseless}, has been studied in~\cite{kalofolias2016learn}. There the graph learning problem is posed as a constrained optimization problem with the constraint set being the set of valid adjacency matrices, and the optimization problem is solved using an iterative primal-dual algorithm. In contrast, our modelling greatly simplifies the solution to simple \emph{rank ordering}. Such a modelling is inspired from~\cite{ghosh2006growing}, where the problem to design edge weights that maximize the algebraic connectivity of the graph has been addressed. In \cite{dong2014learning}, the joint graph learning and denoising problem has been addressed, i.e., the problem that we study in Section~\ref{sec:Noisy}. An alternating minimization algorithm is proposed, alternating between graph learning and denoising, where the graph learning optimization problem involves a search over the space of all valid graph Laplacians. On the contrary,
we show that this problem can be solved in one-step and it boils down to the design of a sparse edge sampling function.

Graph topology identification is also investigated in \cite{segarra2016network} under the assumption that the eigenvectors of the graph Laplacian are known, which is a much stronger assumption. Although the eigenvectors can be computed from graph data (or the sample covariance matrix) when it is stationary with respect to the graph~\cite{marques2016stationary,chepuri2016subsampling}, the graph signals need not always be vertex stationary. In any case, the estimated eigenvectors are not error free due to limited data records. In most of the existing approaches~\cite{dong2014learning,kalofolias2016learn,segarra2016network}, graph sparsification is (or can be) achieved by penalizing the $\ell_1$-norm of the graph Laplacian matrix, adjacency matrix or the shift operator, however, there is no explicit handle to control the number of edges, unlike the proposed approach. In a related line of research, \cite{spielman2011spectral,batson2012twice} investigate computing sparse graphs that approximate a given graph spectrally, which means that their Laplacian matrices have similar quadratic forms.

\section{Problem setup}

Consider a dataset with $N$ real valued elements, which are defined on the vertices of an undirected graph $\mathcal{G} = (\mathcal{V},\mathcal{E})$, where the vertex set $\mathcal{V} = \{v_1,\cdots,v_N\}$ denotes the set of nodes, and the edge set $\mathcal{E}$ reveals the connection between the nodes. We refer to such datasets as {\it graph signals}.  We assume that the length of the graph signals (thus the number of nodes), i.e., $N$ is known. However, the edge set is not known. Therefore, we assume a \emph{complete graph} $\mathcal{G}(\mathcal{V},\mathcal{E})$ as a candidate graph in which each node is connected to every other node with the number of edges $|\mathcal{E}| = M = N(N-1)/2$, and aim to determine a subgraph of $\mathcal{G}$ by choosing a subset of edges, $\mathcal{E}_s$, from the edge set $\mathcal{E}$ of this candidate graph.  

Any undirected graph topology is basically determined by its graph Laplacian matrix, which essentially reveals the connectivity of the graph. Let us denote the graph Laplacian matrix (i.e., a symmetric matrix) of the complete graph by ${\boldsymbol L} \in \mathbb{S}^{N}$, where $[{\boldsymbol L}]_{i,j}$ is nonzero, say equal to 1, only if $i=j$ or $(i,j) \in \mathcal{E}$. The symmetric matrix ${\boldsymbol L}$ can be expressed in terms of the so-called \emph{incidence matrix}, ${\boldsymbol A}  = [{\boldsymbol a}_1,\cdots,{\boldsymbol a}_M] \in \mathbb{R}^{N \times M}$ as
\[
{\boldsymbol L} = {\boldsymbol A}{\boldsymbol A}^T = \sum_{m=1}^M  {\boldsymbol a}_m{\boldsymbol a}_m^T,
\]
where the $m$-th column of ${\boldsymbol A}$, i.e., ${\boldsymbol a}_m$ denotes a length-$N$ edge vector with entries $[{\boldsymbol a}_m]_i = 1$, $[{\boldsymbol a}_m]_j= -1$, and it has zeros elsewhere, for an edge $m$ connecting nodes $i$ with $j$ (more generally, ${\boldsymbol a}_m$ is determined only up to a sign). 

Let us now denote the subgraph $\mathcal{G}_s(\mathcal{V},\mathcal{E}_s)$ with the edge set $\mathcal{E}_s \subset \mathcal{E}$ such that $|\mathcal{E}_s| = K \ll M$. We will refer to such a subgraph with $K$ edges as a $K$-sparse graph. We connect such a $K$-sparse graph 
$\mathcal{G}_s$  to ${\boldsymbol L}$ through a \emph{sparse edge selection} vector ${\boldsymbol w} = [w_1,w_2,\cdots,w_M]^T \in \{0,1\}^M$, where $w_m = 1$ if an edge belongs to the edge subset $\mathcal{E}_s$, and $w_m = 0$ otherwise. In terms of  ${\boldsymbol w}$, $|\mathcal{E}_s| = K$ means $\|{\boldsymbol w}\|_0 = K$. (The notation $\|{\boldsymbol w}\|_0$ counts the number of non-zero entries in ${\boldsymbol w}$.)
Finally, we can write the Laplacian matrix of the $K$-sparse graph, ${\boldsymbol L}_s$, as a function of ${\boldsymbol w}$ as 
\begin{equation}
\label{eq:graph_w}
{\boldsymbol L}_s({\boldsymbol w}) = \sum_{m=1}^Mw_m {\boldsymbol a}_m{\boldsymbol a}_m^T.   
\end{equation}

In what follows, we will optimally design the edge sampling function ${\boldsymbol w}$ to recover the graph that sufficiently explains the data.

\section{Learning from Noiseless Graph Signals} \label{sec:Noiseless}

Let ${\boldsymbol x} = [x_1,x_2,\cdots,x_N]^T \in \mathbb{R}^N$ be a graph signal defined on the vertices $\mathcal{V}$ of a graph. 
The smoothness and the spectral content of the signal both depend on the underlying graph topology. The Laplacian quadratic form given by ${\boldsymbol x}^T{\boldsymbol L}_s({\boldsymbol w}){\boldsymbol x}$ quantifies how smooth the graph signal ${\boldsymbol x}$ is with respect to the underlying graph~\cite{shuman2013Emerging}. In particular, the signal ${\boldsymbol x}$ is smoothest with respect to the graph with $K$ edges for low values of ${\boldsymbol x}^T{\boldsymbol L}_s({\boldsymbol w}){\boldsymbol x}$.

\subsection{Problem statement: noiseless setting}
Suppose we are given $L$ graph signals denoted by the vectors $\{{\boldsymbol x}_{k}\}_{k=1}^L$, and they are collected in an $N \times L$ matrix 
${\boldsymbol X} = [{\boldsymbol x}_1,\cdots,{\boldsymbol x}_L]$. We are interested in recovering the graph Laplacian (in other words, the graph topology) under the prior
information that the graph signals are smooth with respect to a $K$-sparse graph. More formally, we state the following.

\begin{myprob}Given the graph signals $\{{\boldsymbol x}_k\}_{k=1}^{L}$, determine a graph with $K$ edges such that the graph signals have smooth variations on the resulting graph.
\end{myprob}

Mathematically, the above problem can be cast as the following optimization problem:
\begin{equation}
\label{eq:graph_learning_1}
\argmin_{{\boldsymbol w} \in \mathcal{W}}  \quad \frac{1}{L}\sum_{k=1}^L {\boldsymbol x}_k^T{\boldsymbol L}_s({\boldsymbol w}){\boldsymbol x}_k  = \frac{1}{L} {\rm tr}\{{\boldsymbol X}^T{\boldsymbol L}_s({\boldsymbol w}){\boldsymbol X}\},
\end{equation}
where 
$
\mathcal{W} = \{{\boldsymbol w} \in \{0,1\}^M \, \vert \, \|{\boldsymbol w}\|_0 = K\}
$ is the constraint set that restricts the number of edges. 

\subsection{Solver}
Problem \eqref{eq:graph_learning_1} is a cardinality constrained Boolean optimization problem, hence nonconvex. 
 By recalling that ${\boldsymbol L}_s({\boldsymbol w}) = \sum_{m=1}^Mw_m {\boldsymbol a}_m{\boldsymbol a}_m^T$, we can express the cost function in  \eqref{eq:graph_learning_1} as a linear function in ${\boldsymbol w}$, i.e., we have 
\begin{equation}
\frac{1}{L} {\rm tr}\left\{{\boldsymbol X}^T{\boldsymbol L}_s({\boldsymbol w}){\boldsymbol X}\right\} =  \sum_{m=1}^M w_m {\rm tr}\left\{{\boldsymbol X}^T ({\boldsymbol a_m}{\boldsymbol a_m}^T){\boldsymbol X}\right\}.
\end{equation}
Introducing the length-$M$ vector ${\boldsymbol c} = [c_1,c_2,\ldots,c_M]^T$ with $c_m = {\rm tr}\left\{{\boldsymbol X}^T ({\boldsymbol a_m}{\boldsymbol a_m}^T){\boldsymbol X}\right\}$, we can write \eqref{eq:graph_learning_1} as 
\begin{equation}
\label{eq:graph_sorting_problem}
\argmin_{{\boldsymbol w} \in \{0,1\}^M}  \quad {\boldsymbol c}^T{\boldsymbol w} \quad {\rm s. to}  \quad \|{\boldsymbol w}\|_0 = K.
\end{equation}
The above Boolean linear programming problem admits an explicit solution and computing the optimal solution is straightforward. It is solved by sorting the entries of ${\boldsymbol c}$ in an ascending ordering. More specifically, the solution ${\boldsymbol w}$ will have entries equal to 1 at indices corresponding to the $K$ smallest entries of ${\boldsymbol c}$, and others are set to zero (ties may be broken arbitrarily). 
Computationally, the sorting algorithm costs $\mathcal{O}(K \log K)$, and with a parallel implementation (e.g., on different processors), the computational complexity will be as low as $\mathcal{O}(K)$~\cite{blum2008energy,chepuri2015sparse}.
We give another interpretation of this result through the following remark.

\begin{myrem}
Let us suppose the graph signal is stochastic with covariance matrix ${\boldsymbol R}_{\boldsymbol x} = \mathbb{E}\{{\boldsymbol x}{\boldsymbol x}^T\} \in \mathbb{R}^{N \times N}$. Then, the solution to~\eqref{eq:graph_sorting_problem} would select $K$ edges between those nodes having the highest cross-correlation, i.e., it will  
add an edge between the $i$th and the $j$th node if the variables $x_i$ and $x_j$ are strongly correlated.  To see this, we express the cost function in \eqref{eq:graph_learning_1} as
\begin{equation*}
\begin{aligned}
 L^{-1}{\rm tr}\{{\boldsymbol X}^T{\boldsymbol L}_s({\boldsymbol w}){\boldsymbol X}\} &=  {\rm tr}\left\{{\boldsymbol L}_s({\boldsymbol w})\widehat{\boldsymbol R}_{\boldsymbol x}\right\} \nonumber\\ &= \sum_{m=1}^M w_m  ({\boldsymbol a_m}^T\widehat{\boldsymbol R}_{\boldsymbol x}{\boldsymbol a}_m)
 \end{aligned}
\label{eq:inverse_cov}
\end{equation*}
where $\widehat{\boldsymbol R}_{\boldsymbol x} =  \frac{1}{L}{\boldsymbol X}{\boldsymbol X}^T \in \mathbb{R}^{N \times N}$ is the sample data covariance matrix.
Recalling the definition of ${\boldsymbol a_m}$, it is easy to see that the term ${\boldsymbol a_m}^T\widehat{\boldsymbol R}_{\boldsymbol x}{\boldsymbol a}_m = [\widehat{\boldsymbol R}_{\boldsymbol x}]_{i,i} + [\widehat{\boldsymbol R}_{\boldsymbol x}]_{j,j} - 2[\widehat{\boldsymbol R}_{\boldsymbol x}]_{i,j}$ is small if the $i$th and $j$th nodes are highly correlated and we have sufficient samples to compute the sample covariance matrix.
\end{myrem} 
By modelling the graph topology through an {edge selection vector}, the graph learning problem can be solved optimally using a simple and elegant solution with a controlled sparsity level, whereas optimizing directly the graph Laplacian~\cite{dong2014learning} or the adjacency matrix~\cite{kalofolias2016learn} leads to a more complicated suboptimal solution with no explicit handle to control the graph sparsity.

\section{Learning from Noisy Graph Signals} \label{sec:Noisy}
In many cases, we might not have access to the true graph signals. Suppose we observe a noisy version of the graph signal,  ${\boldsymbol x}_k$, as
\begin{equation}
\label{eq:observationmodel}
{\boldsymbol y}_k = {\boldsymbol x}_k + {\boldsymbol n}_k \in \mathbb{R}^N,
\end{equation}
and we are given $L$ such observations for $k=1,2,\ldots,L$, where we assume that ${\boldsymbol n}_k$ is zero-mean white Gaussian noise of variance $\sigma^2$.  To recover ${\boldsymbol x}_k$ based on the smoothness assumption, typically a least-squares problem is solved with a \emph{Tikhonov regularization}, ${\boldsymbol x}_k^T{\boldsymbol L}{\boldsymbol x}_k$, to enforce the prior information that the noiseless graph signal ${\boldsymbol x}_k$ is smooth with respect to the underlying graph. More specifically, the following optimization problem (assuming, for a moment that the graph, i.e., ${\boldsymbol w}$ is known) is solved~\cite{shuman2013Emerging}:
\begin{equation}
\label{eq:graphdenois}
\argmin_{\{{\boldsymbol x}_k\}_{k=1}^L} \quad  \frac{1}{L}\sum_{k=1}^L \left( \|{\boldsymbol y}_k - {\boldsymbol x}_k\|_2^2 + \gamma {\boldsymbol x}_k^T{\boldsymbol L}_s({\boldsymbol w}){\boldsymbol x}_k \right),
\end{equation}
where the regularization parameter $\gamma > 0$ controls the amount of smoothness. This graph denoising problem has an explicit solution given by
\[
\widehat{\boldsymbol x}_k = [{\boldsymbol I} + \gamma {\boldsymbol L}_s({\boldsymbol w})]^{-1}{\boldsymbol y}_k, \, k=1,\cdots,L.
\]
\vskip-1mm
\subsection{Problem statement: noisy setting}
\vskip-1mm
Having given the above denoising inference problem at hand, we will now formally state the problem of interest. 

\begin{myprob}Given the observations $\{{\boldsymbol y}_k\}_{k=1}^{L}$ that is related to the unknown graph signal ${\boldsymbol x}_k$ as in \eqref{eq:observationmodel}, determine the $K$-sparse graph such that the estimate $\widehat{\boldsymbol x}_k$ has the lowest possible estimation error, and it is smooth with respect to the recovered graph.
\end{myprob}

Sparse graph learning for the denoising inference problem can be mathematically formulated as follows: 
\begin{equation}
\label{eq:learning_prob}
\argmin_{\{{\boldsymbol x}_k\}_{k=1}^L, {\boldsymbol w} \in \mathcal{W}} \, \frac{1}{L}\sum_{k=1}^L ( \|{\boldsymbol y}_k - {\boldsymbol x}_k\|_2^2 + \gamma \, {\boldsymbol x}_k^T{\boldsymbol L}_s({\boldsymbol w}){\boldsymbol x}_k)
\end{equation}
whose solution is denoted as $(\{\widehat{\boldsymbol x}_k\}_{k=1}^L, \widehat{\boldsymbol w})$.
This formulation is different from~\cite{dong2014learning}, as \cite{dong2014learning}  solves an optimization problem over the
space of all possible graph Laplacians (instead of parameterizing the graph with ${\boldsymbol w} \in \mathcal{W}$) without sparsifying the graph. It can nevertheless be done through an extra $\ell_1$-norm penalty term.  

The above problem \eqref{eq:learning_prob} is noncovex due to the Boolean and cardinality constraints on ${\boldsymbol w}$ and the coupling between the optimization variables in the second term of \eqref{eq:learning_prob}. We provide two methods to solve it. The first one is a straightforward approach based on alternating descent, while the second one is based on convex relaxation.
\subsection{Alternating minimization} \label{sec:am}
The optimization problem \eqref{eq:learning_prob} can be solved using alternating minimization with respect to $\{{\boldsymbol x}_k\}_{k=1}^L$ and
${\boldsymbol w}$. That is,  given ${\boldsymbol w}$, the problem in \eqref{eq:learning_prob} reduces to a linear system in the unknown ${\boldsymbol X}$, which admits a closed form solution; while given $\{{\boldsymbol x}_k\}_{k=1}^L$, it reduces to a Boolean linear programming problem, which admits an analytical solution with respect to ${\boldsymbol w}$ based on rank ordering. These observations suggest an iterative alternating minimization algorithm yielding successive estimates of $\{{\boldsymbol x}_k\}_{k=1}^L$ with fixed ${\boldsymbol w}$, and alternately of ${\boldsymbol w}$ with fixed
$\{{\boldsymbol x}_k\}_{k=1}^L$. Specifically, with the iterate of ${\boldsymbol w}$ given per iteration $i \geq 0$,  i.e., ${\boldsymbol w}[i]$, we solve for ${\boldsymbol X}[i]$ using a matrix inversion as
\[
{\boldsymbol X}[i] = {\boldsymbol X}_{\rm min}({\boldsymbol w}[i]) 
\]
with
\begin{equation}
\label{eq:alt_w}
\begin{aligned}
 {\boldsymbol X}_{\rm min}({\boldsymbol w}) &= \argmin_{{\boldsymbol X}} \,  \|{\boldsymbol Y} - {\boldsymbol X}\|_F^2 + \gamma \, {\rm tr} \{{\boldsymbol X}^T{\boldsymbol L}_s({\boldsymbol w}){\boldsymbol X}\} \\
 &=[{\boldsymbol I} + \gamma {\boldsymbol L}_s({\boldsymbol w})]^{-1}{\boldsymbol Y},
\end{aligned}
\end{equation}
where ${\boldsymbol Y} = [{\boldsymbol y}_1,{\boldsymbol y}_2,\cdots, {\boldsymbol y}_L]$ is the data matrix of size $N \times L$.

Once ${\boldsymbol X}[i]$ is available, ${\boldsymbol w}[i+1]$ can be obtained by solving the Boolean linear program [cf. \eqref{eq:graph_sorting_problem}]
\[
{\boldsymbol w}[i+1] = \argmin_{{\boldsymbol w} \in \{0,1\}^M}  \quad \sum_{m=1}^M w_mc_m[i+1] \quad {\rm s. to}  \quad \|{\boldsymbol w}\|_0 = K,
\]
where $c_m[i+1] = {\rm tr}\left\{{\boldsymbol X}^T[i+1] ({\boldsymbol a_m}{\boldsymbol a_m}^T){\boldsymbol X}[i+1]\right\}$. In spite of the Boolean and cardinality constraints in the above problem, there exists a simple analytical solution for ${\boldsymbol w}[i+1]$ based on sorting $\{c_m[i+1]\}_{m=1}^M$, i.e., the solution ${\boldsymbol w}[i+1]$ will have entries equal to 1 at indices corresponding to the $K$ smallest entries in $\{c_m[i+1]\}_{m=1}^M$ and zeros otherwise. The iterations are initialized at $i = 0$ by randomly generating ${\boldsymbol w}[i+1]$ from a uniform distribution over $\mathcal{W}$. The above alternating minimization method is computationally very attractive, and consists of two simple known solutions per iteration. However, the algorithm converges only to a stationary point of \eqref{eq:learning_prob}, and it suffers from the choice of the initial estimate.  


The algorithm proposed in \cite{dong2014learning} is also along the lines of alternating minimization, except that the graph learning step involves a complicated optimization over the space of all possible valid Laplacian matrices.  
\vskip-3mm
\subsection{Convex relaxation} \label{sec:cvx}
\vskip-1mm
To avoid the issues related to the initialization of the alternating minimization algorithm, in what follows we propose a one-step solution based on convex relaxation. We can rewrite the formulation in \eqref{eq:learning_prob} alternatively as
\begin{equation}
\widehat{\boldsymbol w} = \argmin_{{\boldsymbol w} \in \mathcal{W}} \quad  r({\boldsymbol w});  \quad
\widehat{\boldsymbol X} = {\boldsymbol X}_{\rm min}(\widehat{\boldsymbol w}) \label{eq:learning_prob2}
\end{equation}
with 
\[
r({\boldsymbol w}) = \|{\boldsymbol Y} - {\boldsymbol X}_{\rm min}({\boldsymbol w})\|_F^2 
+ \gamma \, {\rm tr}\{{\boldsymbol X}_{\rm min}^T({\boldsymbol w}){\boldsymbol L}_s({\boldsymbol w}){\boldsymbol X}_{\rm min}({\boldsymbol w})\}
\]
and
\begin{equation}
\label{eq:learning_xmin}
 [{\boldsymbol I} + \gamma {\boldsymbol L}_s({\boldsymbol w})] {\boldsymbol X}_{\rm min}({\boldsymbol w}) ={\boldsymbol Y}.
\end{equation}
The computational complexity of solving the linear system of equations \eqref{eq:learning_xmin} decreases as the sparsity in ${\boldsymbol w}$ increases. Furthermore, the estimates $\{\widehat{\boldsymbol x}_k\}_{k=1}^L$ and $\widehat{\boldsymbol w}$ in \eqref{eq:learning_prob2} are still the same as in \eqref{eq:learning_prob}. 
\begin{figure*}[!t]
\centering
       		\psfrag{celcius}{\footnotesize $^\circ$ C}
		\psfrag{5}{\tiny 5} \psfrag{6}{\tiny 6} \psfrag{7}{\tiny 7} \psfrag{8}{\tiny 8} \psfrag{9}{\tiny 9} \psfrag{10}{\tiny 10} \psfrag{11}{\tiny 11}
\begin{subfigure}[t]{0.4\textwidth}
                \includegraphics[width=\columnwidth, height=1.5in]{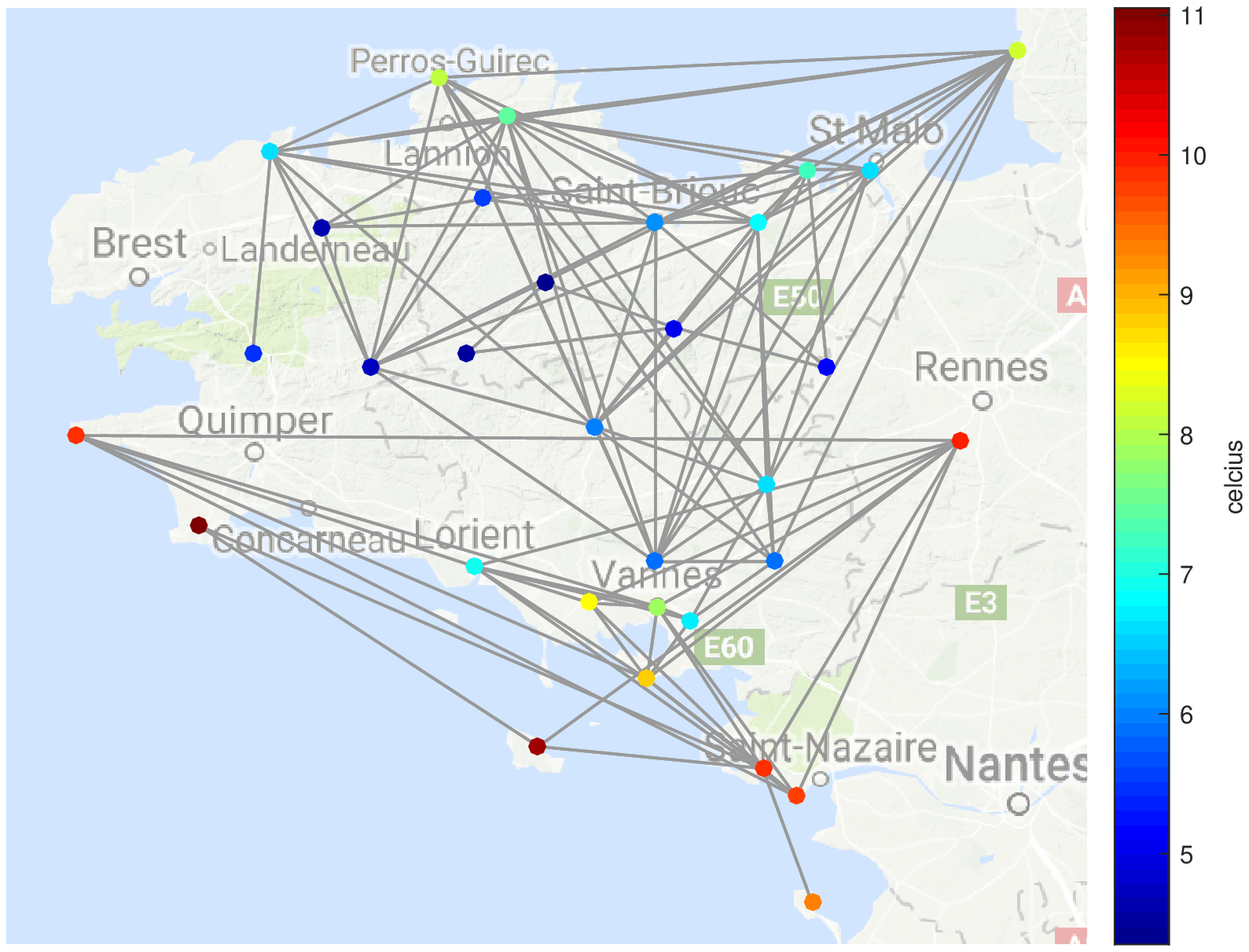}
                \caption{Noiseless setting}
                \label{fig:sparsegraph_noisless}
\end{subfigure}%
~
\begin{subfigure}[t]{0.4\textwidth}
                \includegraphics[width=\columnwidth,height=1.5in]{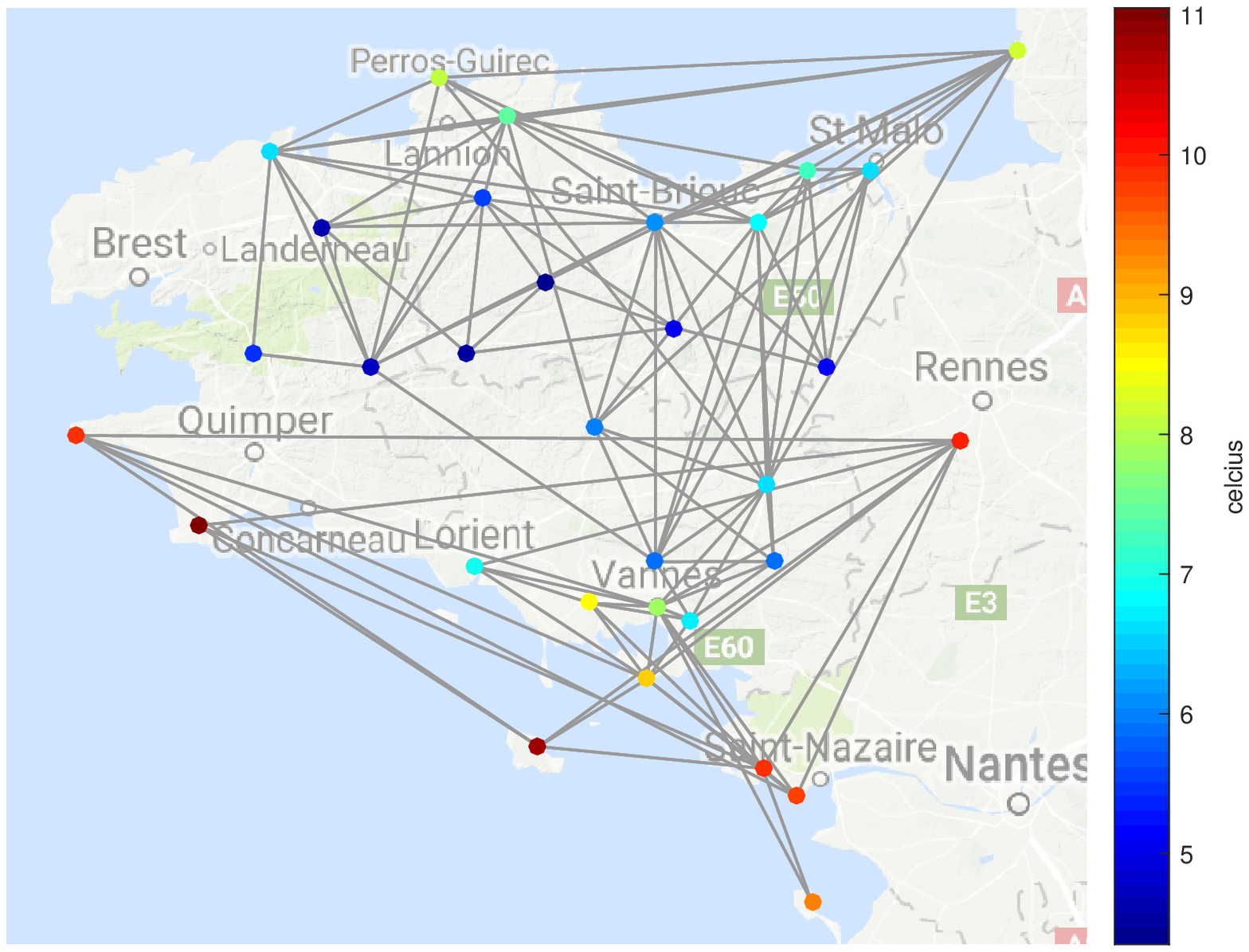}
                \caption{Noisy setting}
                \label{fig:sparsegraph_noisy_am}
\end{subfigure}%
        \caption{\footnotesize{\emph{Sparse graph learning}: The colored dots indicate the temperature values. (a) Noiseless case. Graph with $K=110$ edges recovered by solving \eqref{eq:graph_learning_1}. (b) Noisy case: Convex relaxation is used to recover a graph with $K=110$ edges using \eqref{eq:convexrelax}.}}
        \label{fig:sparsegraph}
\vskip-4mm
\end{figure*}

Plugging the solution to \eqref{eq:learning_xmin} in $r({\boldsymbol w})$ and after some straightforward matrix algebra, we can express the regularized residual squared, $r({\boldsymbol w})$, 
as 
\begin{equation}
\begin{aligned}
r({\boldsymbol w}) 
&={\rm tr}\left\{{\boldsymbol Y}^T [{\boldsymbol I} + \gamma {\boldsymbol L}_s({\boldsymbol w})]^{-1}{\boldsymbol Y}\right\} \\
&\hskip12mm+ \gamma {\rm tr}\left\{{\boldsymbol Y}^T {\boldsymbol L}_s({\boldsymbol w}){\boldsymbol Y}\right\} - \|{\boldsymbol Y}\|_F^2.
\end{aligned}
\end{equation}

Relaxing the cardinality constraint $\|{\boldsymbol w}\|_0 = K$ with ${\boldsymbol 1}^T{\boldsymbol w} = K$ and the Boolean constraints $\{0,1\}^M$ with linear inequality constraints related to the box constraint $[0,1]^M$, the optimization problem  \eqref{eq:learning_prob2} will be convex on ${\boldsymbol w} \in [0,1]^M$. To see this, we introduce a variable
\[
{\boldsymbol Z} = {\boldsymbol Y}^T [{\boldsymbol I} + \gamma {\boldsymbol L}_s({\boldsymbol w})]^{-1}{\boldsymbol Y} + \gamma {\boldsymbol Y}^T {\boldsymbol L}_s({\boldsymbol w}){\boldsymbol Y} \in \mathbb{R}^{L \times L}
\]
and obtain a semidefinite program:
\begin{equation}
\begin{aligned}
\label{eq:convexrelax}
&\argmin_{{\boldsymbol Z},{\boldsymbol w}} \quad  {\rm tr}\{{\boldsymbol Z}\}\\
& {\rm s.to} \quad\left[\begin{array}{cc}{\boldsymbol Z} - \gamma {\boldsymbol Y}^T{\boldsymbol L}_s({\boldsymbol w}){\boldsymbol Y} & {\boldsymbol Y}^T \\{\boldsymbol Y} & {\boldsymbol I} + \gamma {\boldsymbol L}_s({\boldsymbol w}) \end{array}\right] \succeq {\boldsymbol 0}_{L+N},
\\
&\hskip10mm{\boldsymbol 1}^T{\boldsymbol w}= K,  \,\, 0 \leq w_m \leq 1, m=1,2,\ldots,M,
\end{aligned}
\end{equation} 
with variables ${\boldsymbol w}$ and ${\boldsymbol Z}$, and recall that ${\boldsymbol L}_s({\boldsymbol w}) = \sum_{m=1}^Mw_m {\boldsymbol a}_m{\boldsymbol a}_m^T$. A standard off-the-shelf solver can be used for solving the semidefinite program in \eqref{eq:convexrelax}. For large-scale 
problems, computationally cheaper first-order (and online) methods for solving \eqref{eq:convexrelax} can be derived as the size of the linear matrix inequality in \eqref{eq:convexrelax} depends on the size of the training data and the number of nodes.
\begin{figure}[tb]
\centering
       		\psfrag{smoothness}{\hskip-3mm\footnotesize${\rm tr}\{{\boldsymbol X}^T{\boldsymbol L}_s({\boldsymbol w}){\boldsymbol X}$\}}
		\psfrag{Noisless prop12345678}{\tiny proposed (noiseless, optimal)}
		\psfrag{Noisless exist12345678}{\tiny primal-dual~\cite{kalofolias2016learn}}
		\psfrag{no of edges}{\footnotesize No. of edges}
\begin{subfigure}[t]{0.8\columnwidth}
                \includegraphics[width=\columnwidth,height=1.5in]{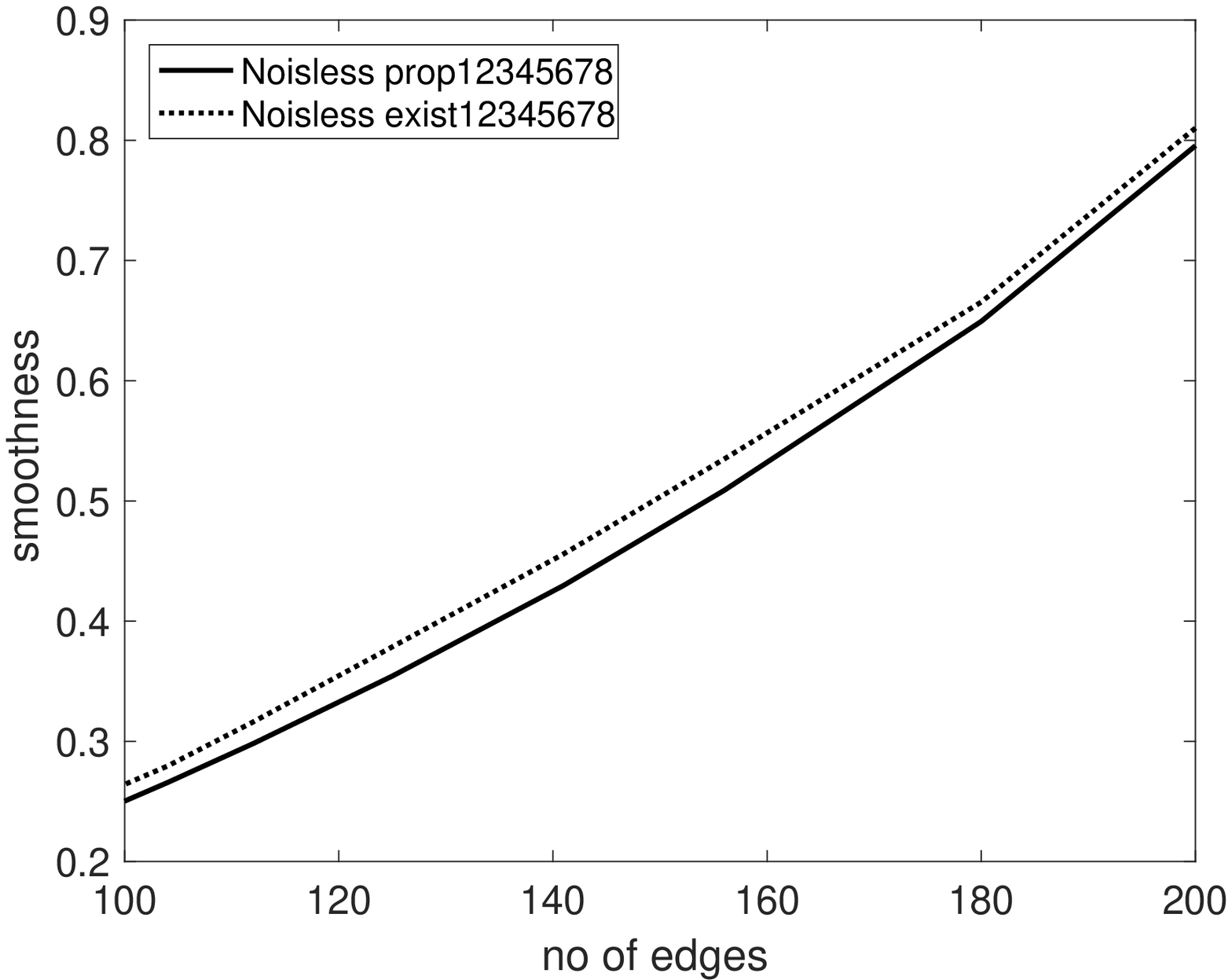}
                \caption{Noiseless setting: smoothness for different graphs}
                \label{fig:smooth_pereval}
\end{subfigure}%
\\[1em]
\psfrag{noise level}{\footnotesize Noise level, $\sigma$}
\psfrag{error}{\hskip-8mm\footnotesize mean squared error}
\psfrag{Noisless prop12345678}{\tiny proposed (noiseless, sorting)}
\psfrag{Noisy am12345678}{\tiny alternating min. in Sec.~\ref{sec:am}}
\psfrag{Noisy exist12345678}{\tiny alternating min. from~\cite{dong2014learning}}
\psfrag{Noisy cvx12345678}{\tiny one-step cvx opt in Sec.~\ref{sec:cvx}}
\begin{subfigure}[t]{0.8\columnwidth}
                \includegraphics[width=\columnwidth, height=1.5in]{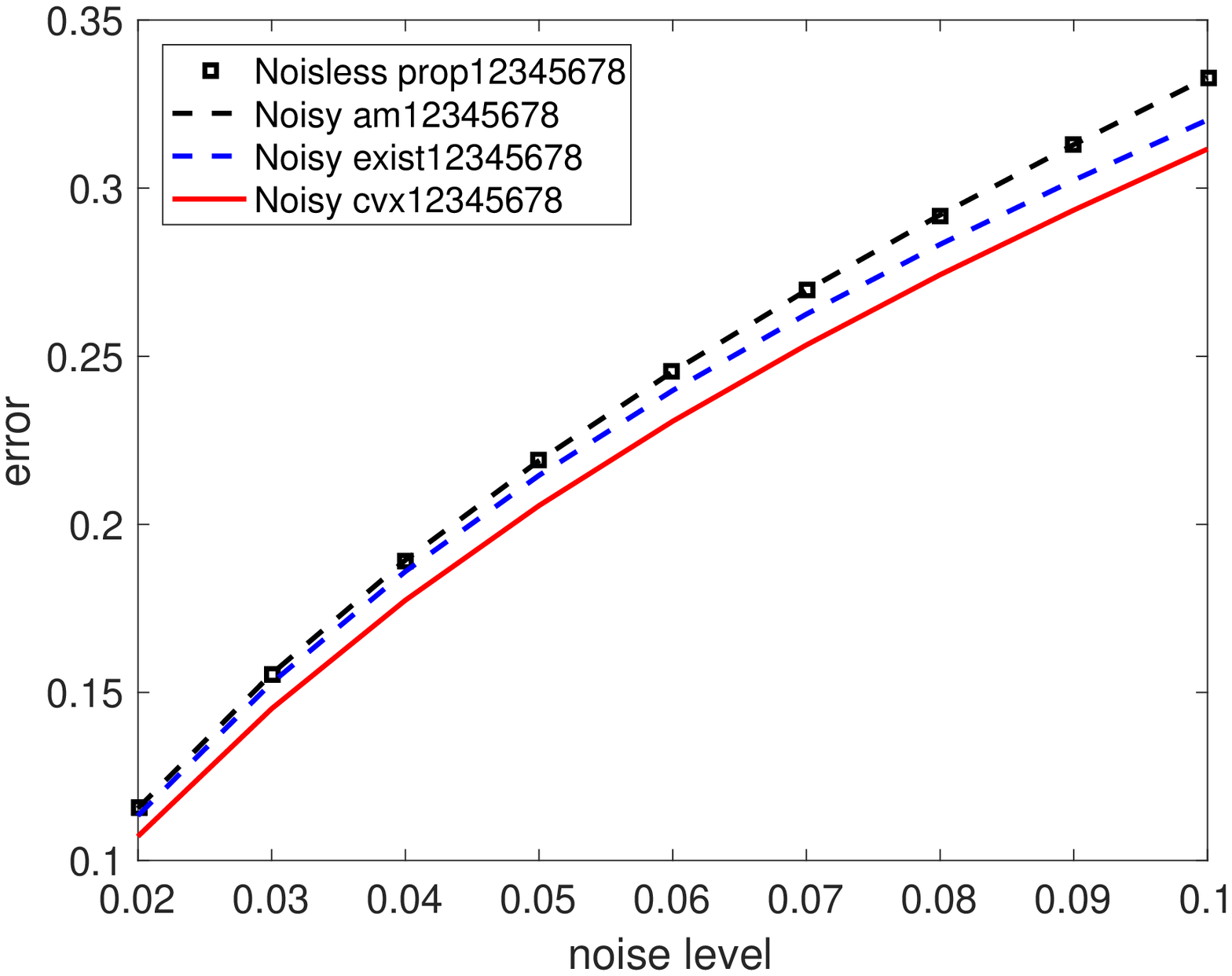}
                \caption{Denoising error for different noise levels}
                \label{fig:mse_pereval}
\end{subfigure}%
        \caption{\footnotesize{Performance evaluation.}}
        \label{fig:per_eval}
\vskip-4mm
\end{figure}
\vskip-2mm
\section{Numerical results}
\vskip-1mm
We use temperature measurements collected across $32$ weather stations in the French region of Brittany and the aim is to learn the graph that explains the observed data; see Fig.~\ref{fig:sparsegraph}. There are $744$ observations per weather station available, out of which we use $L=50$ snapshots as the training set and the remaining ones as the evaluation set. One such observation (i.e., a graph signal) on a graph with $N=32$ nodes is shown in Fig.~\ref{fig:sparsegraph}, where the colored dots indicate different temperature readings. The convex optimization problems are solved using the {\tt CVX} toolbox, which internally calls {\tt SDPT3}~\cite{cvx}\let\thefootnote\relax\footnote{\noindent Software and datasets to produce results of this paper can be downloaded from  {\url{http://cas.et.tudelft.nl/~sundeep/sw/icassp17Graphlearning.zip}}}. The candidate graph with $N=32$ will have $M=496$ edges, from which we aim to learn a subgraph with $K=110$ edges.

To begin with, we consider the noiseless case, where the true graph signal is assumed to be known, and graph learning in this case amounts to solving a sorting problem. As shown in Fig.~\ref{fig:sparsegraph_noisless}, we can see that in the learnt graph with $K=110$ edges, edges are present  between nodes that share similar values. Although the proposed approach doesn't always (e.g., for low values of $K$) ensure a well-connected graph, it clusters entities (or correlated nodes) with similar values. Fig.~\ref{fig:smooth_pereval} shows that the cost (i.e., smoothness) of the proposed closed-form sorting solution, which is optimal, is lower than the existing iterative solution~\cite{kalofolias2016learn}.

Next, we consider the noisy setting with the same training data as before, where we perform joint graph learning and denoising. In Fig.~\ref{fig:sparsegraph_noisy_am}, we show the learnt graph with $K=110$ edges based on the convex relaxation approach explained in Sec.~\ref{sec:cvx}. In Fig.~\ref{fig:mse_pereval}, we evaluate the denoising performance based on the learnt graph using the evaluation set. In particular, we show the mean squared error for different values of the noise level, where the mean squared error is computed from $1000$ independent Monte Carlo experiments.
The one-step solution based on convex optimization (cf. Sec.~\ref{sec:cvx}) leads to a lower error as compared to the alternating minimization approaches, which in general converge only to a stationary point. This also holds for our method developed in Sec.~\ref{sec:am}, however, we stress the fact that the proposed alternating minimization (cf. Sec.~\ref{sec:am}) is computationally much less expensive (involving two simple known solutions per iteration) as compared to the iterative solution in~\cite{dong2014learning}. The graph learnt under the noiseless setting does not perform well for denoising. Nevertheless, due its simple solution, it can be used to generate a base graph, which can be further refined for specific graph inference problems.   

\vskip-4mm
\section{Conclusions}
\vskip-2mm
We have studied the problem of learning a sparse graph that adequately explains the data under a smoothness prior. We model the graph learning problem as the design of a sparse edge sampling function. In other words, we express the graph Laplacian in terms of an edge selection vector. We have considered both the noiseless and noisy setting. In the noiseless setting, designing the edge selection vector is elegant, and it boils down to a simple low-complexity sorting problem. However, in the presence of noise, we propose a computationally cheap alternating minimization algorithm as well as a one-step convex relaxation based solution. 
%
%
\pagebreak
\bibliographystyle{IEEEtran}
\bibliography{IEEEabrv,//users/localadmin/Dropbox/Bibfiles/refs,//users/localadmin/Dropbox/Bibfiles/strings}

\begin{thebibliography}{10}
\providecommand{\url}[1]{#1}
\csname url@samestyle\endcsname
\providecommand{\newblock}{\relax}
\providecommand{\bibinfo}[2]{#2}
\providecommand{\BIBentrySTDinterwordspacing}{\spaceskip=0pt\relax}
\providecommand{\BIBentryALTinterwordstretchfactor}{4}
\providecommand{\BIBentryALTinterwordspacing}{\spaceskip=\fontdimen2\font plus
\BIBentryALTinterwordstretchfactor\fontdimen3\font minus
  \fontdimen4\font\relax}
\providecommand{\BIBforeignlanguage}[2]{{%
\expandafter\ifx\csname l@#1\endcsname\relax
\typeout{** WARNING: IEEEtran.bst: No hyphenation pattern has been}%
\typeout{** loaded for the language `#1'. Using the pattern for}%
\typeout{** the default language instead.}%
\else
\language=\csname l@#1\endcsname
\fi
#2}}
\providecommand{\BIBdecl}{\relax}
\BIBdecl

\bibitem{shuman2013Emerging}
D.~I. Shuman, S.~K. Narang, P.~Frossard, A.~Ortega, and P.~Vandergheynst, ``The
  emerging field of signal processing on graphs: Extending high-dimensional
  data analysis to networks and other irregular domains,'' \emph{{IEEE} Signal
  Process. Mag.}, vol.~30, no.~3, pp. 83--98, 2013.

\bibitem{sandryhaila2014big}
A.~Sandryhaila and J.~M. Moura, ``Big data analysis with signal processing on
  graphs: Representation and processing of massive data sets with irregular
  structure,'' \emph{{IEEE} Signal Process. Mag.}, vol.~31, no.~5, pp. 80--90,
  2014.

\bibitem{slavakis2014modeling}
K.~Slavakis, G.~Giannakis, and G.~Mateos, ``Modeling and optimization for big
  data analytics:(statistical) learning tools for our era of data deluge,''
  \emph{{IEEE} Signal Process. Mag.}, vol.~31, no.~5, pp. 18--31, 2014.

\bibitem{dong2014learning}
X.~Dong, D.~Thanou, P.~Frossard, and P.~Vandergheynst, ``Learning laplacian
  matrix in smooth graph signal representations,'' \emph{arXiv preprint
  arXiv:1406.7842}, 2014.

\bibitem{kalofolias2016learn}
V.~Kalofolias, ``How to learn a graph from smooth signals,'' in
  \emph{Proceedings of the 19th International Conference on Artificial
  Intelligence and Statistics}, 2016, pp. 920--929.

\bibitem{ghosh2006growing}
A.~Ghosh and S.~Boyd, ``Growing well-connected graphs,'' in \emph{Proceedings
  of the 45th IEEE Conference on Decision and Control}, 2006, pp. 6605--6611.

\bibitem{segarra2016network}
S.~Segarra, A.~G. Marques, G.~Mateos, and A.~Ribeiro, ``Network topology
  identification from spectral templates,'' \emph{arXiv preprint
  arXiv:1604.02610}, 2016.

\bibitem{marques2016stationary}
A.~Marques, S.~Segarra, G.~Leus, and A.~Ribeiro, ``Stationary graph processes
  and spectral estimation,'' \emph{arXiv preprint arXiv:1603.04667}, 2016.

\bibitem{chepuri2016subsampling}
S.~P. Chepuri and G.~Leus, ``Subsampling for graph power spectrum estimation,''
  \emph{arXiv preprint arXiv:1603.03697}, 2016.

\bibitem{spielman2011spectral}
D.~A. Spielman and S.-H. Teng, ``Spectral sparsification of graphs,''
  \emph{SIAM Journal on Computing}, vol.~40, no.~4, pp. 981--1025, 2011.

\bibitem{batson2012twice}
J.~Batson, D.~A. Spielman, and N.~Srivastava, ``Twice-ramanujan sparsifiers,''
  \emph{SIAM Journal on Computing}, vol.~41, no.~6, pp. 1704--1721, 2012.

\bibitem{blum2008energy}
R.~S. Blum and B.~M. Sadler, ``Energy efficient signal detection in sensor
  networks using ordered transmissions,'' \emph{{IEEE} Trans. Signal Process.},
  vol.~56, no.~7, pp. 3229--3235, 2008.

\bibitem{chepuri2015sparse}
S.~P. Chepuri and G.~Leus, ``Sparse sensing for distributed detection,''
  \emph{{IEEE} Trans. Signal Process.}, vol.~64, no.~6, pp. 1446--1460, 2015.

\bibitem{cvx}
M.~Grant and S.~Boyd, ``{CVX}: Matlab software for disciplined convex
  programming, version 2.0 beta,'' \url{http://cvxr.com/cvx}, Sep. 2012.

\end{thebibliography}

\end{document}